\documentclass[10pt,twocolumn,letterpaper]{article}
\usepackage[accsupp]{axessibility}
\usepackage{cvpr}    
\usepackage{graphicx}
\usepackage{amsmath}
\usepackage{amssymb}
\usepackage{booktabs}
\usepackage{multirow}
\usepackage{makecell}
\usepackage{comment}
\usepackage{amsfonts}
\usepackage{epsfig}
\usepackage{diagbox}
\usepackage[pagebackref,breaklinks,colorlinks]{hyperref}
\usepackage{enumitem}
\usepackage[capitalize]{cleveref}
\crefname{section}{Sec.}{Secs.}
\Crefname{section}{Section}{Sections}
\Crefname{table}{Table}{Tables}
\crefname{table}{Tab.}{Tabs.}

\newcommand{\BEST}[1]{\textbf{\textcolor[rgb]{1.00,0.00,0.00}{#1}}}
\newcommand{\SBEST}[1]{\textbf{\textcolor[rgb]{0.00,0.00,1.00}{#1}}}
\def\cvprPaperID{7237} % *** Enter the CVPR Paper ID here
\def\confName{CVPR}
\def\confYear{2023}

\begin{document}

\title{The Treasure Beneath Multiple Annotations: An Uncertainty-aware Edge Detector}

\author{Caixia Zhou$^{1}$, Yaping Huang$^{1}$\thanks{Corresponding author.}, Mengyang Pu$^{2}$, Qingji Guan$^{1}$, Li Huang$^{1}$, Haibin Ling$^{3}$\\ 
$^1$Beijing Key Laboratory of Traffic Data Analysis and Mining, Beijing Jiaotong University, China\\
$^2$ School of Control and Computer Engineering, North China Electric Power University, China\\
$^3$Department of Computer Science, Stony Brook University, USA\\
{\tt\small \{cxzhou,yphuang,qjguan,20112044\}@bjtu.edu.cn, mengyang.pu@ncepu.edu.cn, hling@cs.stonybrook.edu}
}

\maketitle

\begin{abstract}
  Deep learning-based edge detectors heavily rely on pixel-wise labels which are often provided by multiple annotators. Existing methods fuse multiple annotations using a simple voting process, ignoring the inherent ambiguity of edges and labeling bias of annotators. In this paper, we propose a novel uncertainty-aware edge detector (UAED), which employs uncertainty to investigate the subjectivity and ambiguity of diverse annotations. Specifically, we first convert the deterministic label space into a learnable Gaussian distribution, whose variance measures the degree of ambiguity among different annotations. Then we regard the learned variance as the estimated uncertainty of the predicted edge maps, and pixels with higher uncertainty are likely to be hard samples for edge detection. Therefore we design an adaptive weighting loss to emphasize the learning from those pixels with high uncertainty, which helps the network to gradually concentrate on the important pixels. UAED can be combined with various encoder-decoder backbones, and the extensive experiments demonstrate that UAED achieves superior performance consistently across multiple edge detection benchmarks. The source code is available at \url{https://github.com/ZhouCX117/UAED}.

\end{abstract}

%%%%%%%%% BODY TEXT
\section{Introduction}
\begin{figure}[ht]
 \centering
 \includegraphics[scale=0.57]{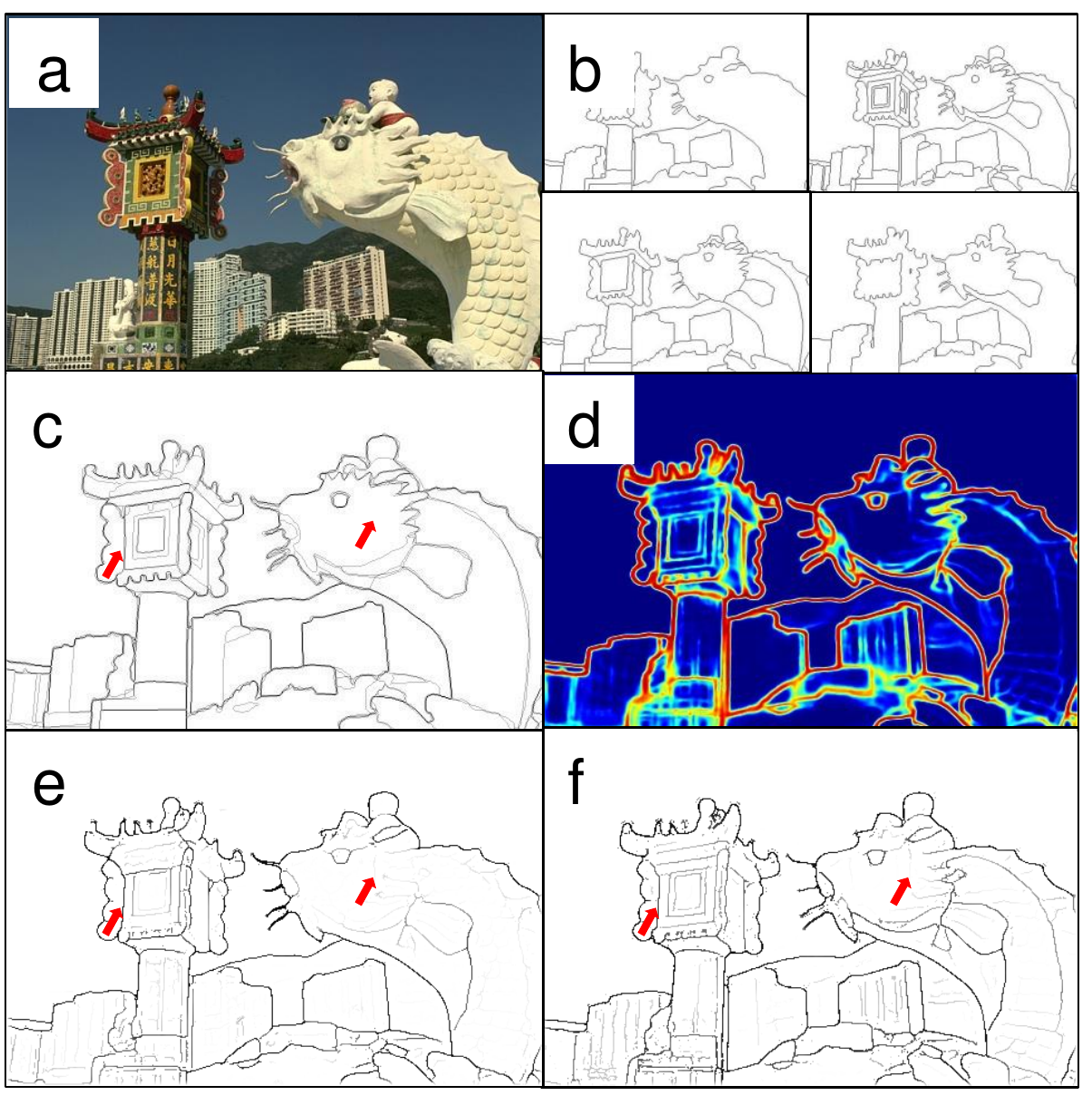} 
 \caption{Illustration of the proposed Uncertainty-Aware Edge Detector (UAED). The first row shows (a) an image from the BSDS test set and (b) four diverse labels by different annotators. The second row shows (c) the final edge label computed by majority voting and (d) our estimated uncertainty map (red means high uncertainty and blue means low uncertainty). The third row shows the edge detection results by (e) EDTER~\cite{pu2022edter} and (f) our UAED, both processed by non-maximum suppression.}
  \vspace{-12pt}
 \label{Fig1}
\end{figure}

Edge detection is a fundamental low-level vision task. It greatly reduces irrelevant information and retains the most important structural attributes. An efficient edge detector can generate structural edges that depict important areas from a whole image, thereby benefiting many downstream tasks~\cite{qin2019basnet,yu2021coupled,liu2020dynamic,nazeri2019edgeconnect,song2018edgestereo}.
Early pioneering methods~\cite{kittler1983accuracy,canny1986computational} compute the gradient and choose suitable thresholds to select pixels with obvious brightness changes. Hand-crafted feature based methods~\cite{martin2004learning,arbelaez2010contour} extract features from low-level cues including density and texture, and then design complex rules to distinguish edges. Benefiting from
the powerful feature representation of Convolution Neural Network (CNN) and Transformer, 
% recent works~\cite{xie2015holistically,liu2017richer,he2019bi,pu2022edter} concentrate on designing better network architectures to learn high-level semantic representations. 
recent works~\cite{xie2015holistically,liu2017richer,he2019bi,pu2022edter} concentrate on designing elaborate network architectures to learn high-level semantic representations.
% With the major breakthrough of  Transformer~\cite{vaswani2017attention} in vision~\cite{dosovitskiy2020image}, EDTER~\cite{pu2022edter} successfully designs a transformer-based edge detector, which raises performance to a new level. %\ling{[maybe shrink this paragraph and merge the first paragraphs?]}

The previous efforts are mainly dedicated to designing advanced networks to extract distinctive features.  
Except for the well-designed models, precise pixel-level annotation is another key factor in building an efficient edge detector under the supervised setting. Due to the complexity of the scenes and the ambiguity of the edges, most of the works~\cite{arbelaez2010contour,mely2016systematic} involve multiple annotators for labeling edges. However, the subjectivity of the annotators, \eg, different people may perceive the same scene differently and annotate the edges at different granularities, leading to inconsistent annotations (Fig.~\ref{Fig1}(b)). Previous methods simply utilize the majority voting strategy to fuse multiple annotations into single ground truth, where all annotations are averaged to generate an edge probability map (Fig.~\ref{Fig1}(c)), ranging from 0 to 1. During training, the pixels with probability higher than a fixed threshold are regarded as positive and the pixels with probability equal to 0 as negative. 
% The pixels with probability lower than the threshold \textcolor{red}{and higher than 0} are dropped.
And the remaining pixels are dropped.
%\textcolor{red}{Such voting}
Such a simple voting process 
neglects the inherent ambiguity and label bias caused by the labeling process.

To address the issues, in this paper, we propose a novel uncertainty-aware edge detection (UAED) framework that converts the deterministic labels into distributions to explore the inherent label ambiguity in the edge detection task. Unlike previous works that focus on architecture modification, we target modeling the uncertainty underlying the multiple edge annotations.

Specifically, the proposed UAED is designed based on the encoder-decoder architecture, where the encoder generates the feature representations followed by two separate decoders. Instead of using fixed labels, we treat the prediction as a learnable Gaussian distribution, whose mean and variance are learned by two decoders respectively, and the variance can be supervised by multiple annotations. 
The learned variance can be naturally regarded as uncertainty, which measures the label ambiguity. Therefore we further utilize the learned uncertainty to boost the performance. Fig.~\ref{Fig1}(d) shows the estimated uncertainty map. {We can observe that the uncertainties of pixels that are close to edges are much higher than those of smooth regions. This phenomenon suggests that pixels with higher uncertainty are visually more important than pixels with lower uncertainty and can be regarded as hard samples for detecting edges.} Thus inspired, unlike most uncertainty estimation methods that regard the pixels with higher uncertainty as unreliable and discard them, we encourage the model to learn more from the hard samples with higher uncertainty progressively. The experiments on two popular edge detection datasets with multiple annotations show the effectiveness of our proposed method. {Compared with transformer-based EDTER~\cite{pu2022edter} (Fig.~\ref{Fig1}(e)), our proposed UAED combined with CNN-based architecture can generate more detailed edges (Fig.~\ref{Fig1}(f)), while requires less computation resource and time.} Our contributions can be summarized as follows:
\begin{itemize}[itemsep=0pt,topsep=0pt,parsep=0pt]
    \item We propose an uncertainty-aware edge detector, named UAED, which captures the inherent ambiguity caused by multiple subjective annotations. To our best knowledge, this is the first work that provides an uncertainty perspective in  edge detection.
    \item We concentrate on the pixels with higher uncertainty that play a more important role in  edge detection, and further design an adaptive weighting loss to emphasize the training from those hard pixels.
    \item UAED can be combined with various encoder-decoder backbones without increasing {much} computation burden. We conduct comprehensive experiments on popular datasets across different model architectures and achieve consistent improvement.
\end{itemize}

\section{Related Work}

% Our work is related to edge detection and uncertainty in deep learning. Next, we will introduce these two parts in turn for a better understanding of our motivation.

\subsection{Edge Detection}
Edge detection is an important vision task have been attracting a great amount of study.  Early methods, such as Sobel~\cite{kittler1983accuracy} and Canny~\cite{canny1986computational}, calculate the gradient of density, color or texture of the images for edge clues. Traditional learning-based methods aim to design hand-crafted features from low-level density and texture cues to train a edge classifier.
For example, Pb~\cite{martin2004learning} defines the changes in brightness, color and texture as the edge features.
gPb~\cite{arbelaez2010contour} utilizes standard Normalized Cuts for detecting edges. 

Benefiting from the success of deep learning technologies, CNN-based methods become predominant edge detectors. Early methods~\cite{shen2015deepcontour,bertasius2015deepedge} are based on image patches that extract features from the predefined patches, and determine whether there are edge pixels. Later, pixel-based models have achieved promising performance. HED~\cite{xie2015holistically} utilizes VGG16~\cite{simonyan2014very} as the backbone and obtains five stage feature maps as the side outputs, which are then fused into final outputs by learnable image-level weights. RCF~\cite{liu2017richer} connects each convolution layer in VGG16 to a convolution layer, and then accumulate the results to attain hybrid features to fully use multi-scale multi-level information. LPCB~\cite{deng2018learning} applies VGG16 as the backbone and utilizes the ResNetXt~\cite{he2016deep} block and deconv module to fuse features across stages to decode the edge maps. Instead of treating side outputs and final outputs the same, BDCN~\cite{he2019bi} approximates the specific edge ground truth for different scales. RINDNet~\cite{pu2021rindnet} uses multi-branch strategy for fine grained edge detection. Recently, a transformer-based edge detector EDTER~\cite{pu2022edter} is proposed, which first splits the input image into a sequence of $16\times 16$ patches to extract global features, and then extracts the short-range local cues on $8\times 8$ patches.

\begin{figure*}[!t]
 \centering
 \includegraphics[scale=0.53]{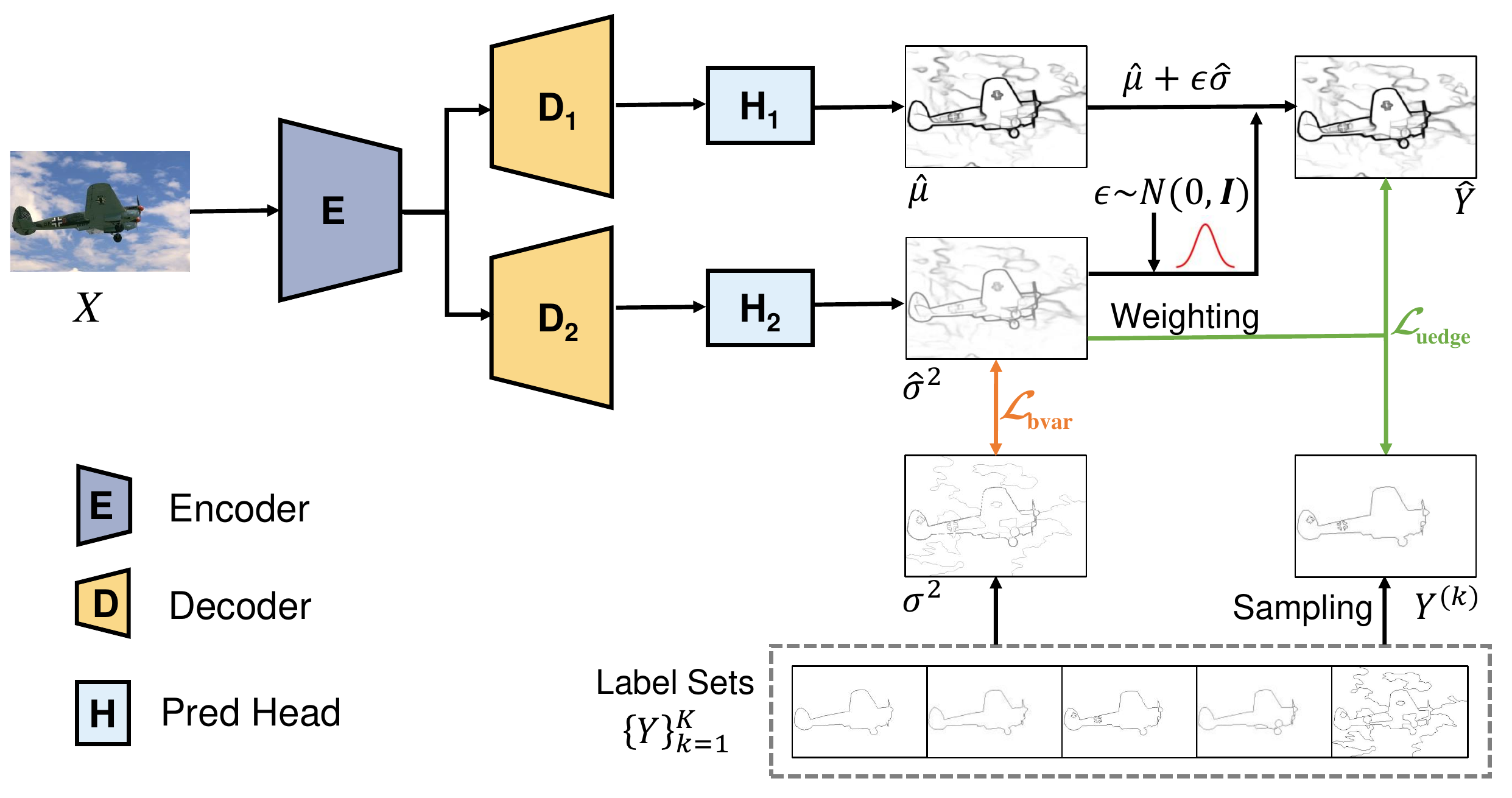}
 \caption{The overall framework of our proposed UAED. Given an input $X$, the encoder $\mathcal{E}$ extracts multi-scale features, which are then fed to two independent decoders $\mathcal{D}_1,\mathcal{D}_2$ and prediction heads $\mathcal{H}_1,\mathcal{H}_2$ to obtain respectively mean $\hat{\mu}$ and variance $\hat{\sigma}^2$. We construct a multi-variate Gaussian distribution according to the learned mean and variance, and sample prediction $\hat{Y}$  from this distribution. The learned variance is supervised by the label variance $\sigma^2$ computed from the labeling sets $\{Y^{(k)}\}_{k=1}^{K}$, and the prediction $\hat{Y}$ is supervised by the label $Y^{(k)}$ sampling from the labeling sets. 
 %\ling{[(1) Why is there a link from $\sigma_p^2$ to $\mathcal{L}_{uedge}$? (2) Adding variables $X,F$, etc. (3) improve the layout.]}
 }
 \vspace{-15pt}
 \label{Fig2}
\end{figure*}

Some other works aim to construct lightweight models~\cite{su2021pixel,deng2021learning} consuming as few resources as possible while maintaining performance. PiDiNet~\cite{su2021pixel} introduces pixel difference convolution (PDC) which integrates the traditional edge detection operators such as Sobel and local binary patterns (LBP) into the popular convolution operations. LDC~\cite{deng2021learning} is an encoder-decoder structure, where the encoder is tested on three different lightweight models, \ie, SqueezeNet~\cite{iandola2016squeezenet}, MobileNetV2~\cite{howard2018inverted}, and  RegNetX~\cite{radosavovic2020designing}. The decoder module designs a spatially Squeeze-and-extraction (SSE) module to explore global contextual information and an SE module to extract local context.

Despite the thorough exploration of the network design and great performance achievement, existing works ignore the exploration of the inherent ambiguity in the label space, which is the focus of our work.

\subsection{Uncertainty in Deep Learning}
The major uncertainty encountered in deep learning~\cite{kendall2017uncertainties} incorporates aleatoric (data) and epistemic (model) uncertainty, where data uncertainty is intrinsic while model uncertainty can be reduced by more data. Common methods for modeling uncertainty involve generative models~\cite{zhang2021dense, Kohl2018probabilistic20,gustafsson2020energy}, introducing new branches~\cite{chang2020data, patro2019u}, and regarding model parameters as a distribution~\cite{gal2016dropout, lakshminarayanan2017simple}.

The techniques based on generative models can be classified into generative adversarial network (GAN)~\cite{zhang2021dense}, energy-based model (EBM)~\cite{gustafsson2020energy,zhang2021learning}, variational autoencoder (VAE)~\cite{Kohl2018probabilistic20,Baumgartner2019PhiSeg21,Zhang2020UCnet13}, normalizing flow (NF)~\cite{kumar2019videoflow,wehrbein2021probabilistic}, and a hybrid of the above methods~\cite{selvan2020uncertainty,valiuddin2021improving}. Introducing additional branches can operate on both the feature space~\cite{chang2020data} and label space~\cite{patro2019u} to convert the deterministic result into a distribution, such as Gaussian and Laplace distribution.
Compared with other methods, modeling uncertainty on the parameters space is relatively 
% time-consuming
time or memory consuming, such as MC-Dropout~\cite{gal2016dropout} and deep ensemble~\cite{lakshminarayanan2017simple}.  

Despite effective uncertainty estimation methods having been successfully applied in many computer vision tasks, they have not been explored on the edge detection task. Our work is the first to model uncertainty in edge detection and design an efficient uncertainty estimation to explore the uncertainty underlying multiple edge labels.

\section{Uncertainty-aware Edge Detection}

The overview of the proposed uncertainty-aware edge detection (UAED) is shown in Fig.~\ref{Fig2}. {Given an image $X \in \mathbb{R}^{H \times W \times 3}$ and its corresponding annotations
$\{Y^{(k)}\}_{k=1}^{K}$, where $Y^{(k)}\in \mathbb{R}^{H \times W \times 1}$ is the $k\textrm{-th}$ annotation and $K$ is the number of annotations. We first feed $X$ into an encoder ($\mathcal{E}$) to extract multi-scale features, and then the extracted features are fed into two independent decoders ($\mathcal{D}_1,\mathcal{D}_2$) and prediction heads ($\mathcal{H}_1,\mathcal{H}_2$) to obtain mean ($\hat{\mu}$) and variance ($\hat{\sigma}^2$) for the learned Gaussian distribution respectively. The learned variance ($\hat{\sigma}^2$) is supervised by the variance computed from the label sets ($\sigma ^2$).
Finally, the sampling from the distribution is regarded as the prediction ($\hat{Y}$), which is supervised by the annotation $Y^{(k)}$ randomly sampled from the label sets.}

{We will detail the edge detection network in Section~\ref{net}, and then introduce the proposed uncertainty estimation in Section~\ref{ue} and the optimized objective in Section~\ref{obj}.}

\subsection{Edge Detection Network}\label{net}
We design our edge detector based on the encoder-decoder architecture, which is widely used in the edge detection task~\cite{deng2021learning,deng2018learning,deng2020deep}. 
Here, we adopt EfficientNet~\cite{tan2019efficientnet} as the encoder and UNet++~\cite{zhou2018unet++} as the decoder.
% considering its state-of-the-art performance. 
 It should be noted that the edge detection network itself is not our contribution. So we do not pay more attention to the design of the network and just borrow the existing excellent framework for convenience. In fact, our proposed uncertainty-driven method can be combined with other edge detection networks, as demonstrated in the experiments in Section~\ref{arch}.

The encoder of our edge detection network consists of eight stages. Unlike the original EfficientNet starting with a convolutional layer that aims to change the input channels, we modify the stride of the first stage so that it can downsample the feature map by half. The seven middle stages are the same as the original EfficientNet, where each stage has different number of convolutional blocks consisting of the convolutional layer, batch normalization layer~\cite{ioffe2015batch} and swish activation layer~\cite{ramachandran2017swish}. We remove the last ninth stage of the original EfficientNet~\cite{tan2019efficientnet} that is a classifier head including pooling and fully connected layers, since edge detection needs a fully convolutional network. 
% Specifically, we store the feature maps from the first, the third, the fourth, the sixth, and the eighth as multi-scale features for objects of different sizes that are fed into the following decoders. 
We store the feature maps from the first, the third, the fourth, the sixth, and the eighth as multi-scale features for objects of different sizes, and then feed them into the following decoders.
Note that the feature maps are down-sampled to $1/2$, $1/4$, $1/8$, $1/16$, and $1/32$ of the original images respectively. The details of the encoder are given in supplementary material.

The decoder of our edge detection network generates high-resolution representations from the 
% extracted
received five-level multi-scale features by dense short-range connections and long-range connections, which is based on UNet++~\cite{zhou2018unet++} and can be viewed as four UNets with different sizes. Each UNet is a U-shape architecture including contracting and expansive path. The contracting path is responsible for the reduction of spatial information and the increase of abstract features. The expansive path up-samples the features and connects them with features from the contracting path to recover the images with high semantic information. 

To summarize, for the input image $X$, we first extract the multi-scale features by the designed encoder. Then the extracted features are fed to the decoder and prediction head, which converts the high-dimensional channels of feature maps to one channel by a convolutional layer. Finally, a sigmoid activation function acts on the result to obtain the predicted edge map $\hat{Y}$, which ranges from 0 to 1.

\subsection{Uncertainty Estimation}\label{ue}
To introduce uncertainty into the edge detection task, we convert the deterministic output into a learnable distribution in the label space. Specifically, we add a decoder and a prediction head, which share the same structure but not the same parameters with the original decoder and prediction head, into the edge detection network. Then we feed the extracted multi-scale features into those two independent decoders ($\mathcal{D}_1, \mathcal{D}_2$) and prediction heads ($\mathcal{H}_1, \mathcal{H}_2$). The outputs are denoted as the mean $\hat{\mu}$ and the variance $\hat{\sigma}^2$ of the predicted edge maps: \begin{equation}
    {\hat{\mu}}=\mathcal{H}_1(\mathcal{D}_1(\mathcal{E}(X)), \ \ \hat{\sigma}^2=\mathcal{H}_2(\mathcal{D}_2(\mathcal{E}(X))).
\end{equation}

The predicted mean and variance are naturally built into a multi-variate Gaussian distribution, and the final prediction $\hat{Y}$ is sampled from the distribution by the reparameterization trick~\cite{kingma2015variational}:
\begin{equation}
% \setlength{\abovedisplayskip}{3pt}
% \setlength{\belowdisplayskip}{3pt}
%    \hat{Y}=\mathcal{S}(\hat{\mu}+\epsilon \cdot \hat{\sigma}), \epsilon \sim  \mathcal{N}(0,\text{I}),
    \hat{Y}={\rm sigmoid}(\hat{\mu}+\epsilon \hat{\sigma}), \ \ \epsilon \sim  \mathcal{N}(0,\text{I}).
\end{equation}
%where $\mathcal{S}$ represents the sigmoid function.

\subsection{Network Training}\label{obj}
The optimization process of UAED contains two kinds of loss functions: balanced mean squared error (MSE) loss function for uncertainty estimation and weighted binary cross-entropy (BCE) loss function.

\vspace{-2mm}\paragraph{Balanced MSE loss for uncertainty estimation.} 
In addition to using a single available ground truth label, we hope to make full use of the entire candidate label sets. Thanks to multiple diverse annotations, we can compute the variance of the label sets ($\sigma^2$) as the supervision of the predicted variance ($\hat{\sigma}^2$), which can further be used to indicate the uncertainty estimation for the annotated labels. Therefore, we use $\mathcal{L}_2$ loss for estimating the variance defined as:
\begin{equation}
    \mathcal{L}_{\rm var}=\sum_{j=1}^{HW} (\hat{\sigma}^2_j-\sigma^2_j)^2,
\end{equation}
{where $j$ denotes the $j$-th pixel in the estimated variance and ground truth variance map.}

However, edge pixels are very sparse for no more than $10\%$ edge pixels in an image, so we use an adaptive weighting scheme to balance the loss, similar to ~\cite{hwang2014contour, liu2017richer}. The weight $\alpha$ for the positive samples is calculated as:
\begin{equation}
\alpha=|Y^{(k)}_-|\big/(|Y^{(k)}_-|+|Y^{(k)}_+|), 
\end{equation}
where $|\cdot|$ denotes the number of pixels, and $Y^{(k)}_-$ and $Y^{(k)}_+$ denote the negative and positive samples respectively in $k\textrm{-th}$ annotated edge map. %Accordingly, the weight for negative samples should be $1-\alpha$. 
Since the number of positive pixels is much smaller than that of negative ones, 
the balanced weight makes the model assign a higher weight to edge pixels. Denoting the weight for $Y_j$ as $M_j$, the balanced MSE loss is utilized to calculate the variance loss:  
\begin{equation}
\mathcal{L}_{\rm bvar}=\sum_{j=1}^{HW} M_j (\hat{\sigma}^2_j-\sigma^2_j)^2,
\end{equation}
where
\begin{equation}
M_j = \alpha Y_j + (1-\alpha)(1-Y_j).
\end{equation}

\vspace{-2mm}\paragraph{Uncertainty-driven loss for edge detection.} Edge detection is a binary classification task, therefore the BCE loss is widely used in this task. Due to the imbalanced data, we add a balance weight similar to the variance loss. The balanced BCE loss can be denoted as: 
% \begin{equation}
%\begin{split}&\mathcal{L}_{\rm edge}   = - \sum_{j}W_{j} \big({Y_{j}^{(k)}}{\rm log}(P_{j})  + (1-{Y_{j}^{(k)}}){\rm log}(1-P_{j}) \big),
%\end{split}
\begin{equation}
\begin{aligned}
\mathcal{L}_{\rm edge}   = - \sum_{j=1}^{HW}M_j \Big( & {Y^{(k)}_j}{\rm log}(\hat{Y}_j)  
\\&+ (1-{Y^{(k)}_j}){\rm log}(1-\hat{Y}_j) \Big),
\end{aligned}
\end{equation}
where $\hat{Y}_j$ and $Y_j^{(k)}$ are the $j$-th pixel of the prediction $\hat{Y}$ and label $Y^{(k)}$, respectively. The 
$k$-th label is randomly selected from the available label sets as the supervision signal.

Since the estimated variance ($\hat{\sigma}$) can be regarded as an indicator of the uncertainty of each pixel sample, it can be utilized to further guide the training of edge detector. Naturally, certain samples should be given higher weights while uncertain ones have low importance for training, which has been popularly used in many computer vision tasks~\cite{zhang2021dense}. Unfortunately, in our experiments, such a strategy can not boost the performance and, even worse, lead to an obvious performance drop (see the ablation study in Section~\ref{abstudy}). In fact, as shown in Fig.~\ref{Fig1}(d), we can observe that the pixels with higher uncertainty are usually close to the edges and boundaries in the estimated uncertainty map, which should not be neglected. Instead, those pixels should be prioritized in the training to enforce the model focus on learning from these difficult edges. Inspired by this observation, we design a different weighting strategy, where the pixels with higher uncertainty will be given large weights.

Besides, to prevent the pixels with higher uncertainty from confusing the model in the early training stage, we finally propose a progressive uncertainty-driven weighting strategy, where the pixels with higher uncertainty will be given larger weights progressively. The corresponding loss function is defined as:
\begin{equation}
\mathcal{L}_{\rm uedge} = \sum_{j=1}^{HW} {\rm{exp}}{ (\beta_t \hat{\sigma}_j)} \mathcal{L}_{\rm edge},
\end{equation}
where $\beta_t=t/T$ denotes an adaptive factor, $t$ means the current epoch and $T$ means the total epochs. The final optimization objection is the sum of balanced MSE loss and weighted BCE loss:
\begin{equation}
\mathcal{L}=\mathcal{L}_{\rm uedge}+ \mathcal{L}_{\rm bvar}.
\end{equation}
% where $\gamma$ is a hyperparameter to control the weight of $\mathcal{L}_{\rm bvar}$. In the experiment, we empirically \ling{set $\gamma$ as 1}.

\section{Experiments}
\subsection{Datasets}
We conduct experiments on two popular edge detection datasets, \ie, BSDS500~\cite{arbelaez2010contour} and Multicue~\cite{mely2016systematic}, which contain multiple annotations for each image. 
\begin{table*}[ht]
\setlength{\abovecaptionskip}{0pt}
\caption{Results on the \textbf{BSDS500}~\cite{arbelaez2010contour} testing set. SS is the single-scale testing, MS is the multi-scale testing, and VOC means training with extra PASCAL VOC data. The best two results are denoted as \BEST{red} and \SBEST{blue} respectively, and the same for other tables.}
\centering
\footnotesize
\renewcommand\arraystretch{1}
\renewcommand\tabcolsep{4pt}
\begin{tabular}{c|c|c|ccc|ccc|ccc|ccc}
    \toprule
   \multicolumn{1}{c|}{\multirow{2}*{Method}}&\multicolumn{1}{c|}{\multirow{2}*{Backbone}}&
   \multicolumn{1}{c|}{\multirow{2}*{Pub.'Year}}&
    \multicolumn{3}{c|}{\multirow{1}*{SS}} &\multicolumn{3}{c|}{\multirow{1}*{MS}}&
    \multicolumn{3}{c|}{\multirow{1}*{SS-VOC}}
    &\multicolumn{3}{c}{\multirow{1}*{MS-VOC}}
    \\
    \cline{4-15}
    &&
    &ODS & OIS& AP 
    &ODS & OIS& AP 
    &ODS & OIS& AP 
       &ODS & OIS& AP \\
    \toprule
     Canny~\cite{canny1986computational} &-     & PAMI'86     & 0.611     & 0.676     & 0.520&-&-&-&-&-&-&-&-&- \\
     gPb-UCM~\cite{arbelaez2010contour} &- & PAMI'10     & 0.729     & 0.755     & 0.745&-&-&-&-&-&-&-&-&- \\
    SCG~\cite{xiaofeng2012scg}    &-& NeurIPS'12   & 0.739     & 0.758     & 0.773&-&-&-&-&-&-&-&-&- \\
     SE~\cite{dollar2014fast}   &-& PAMI'14     & 0.743     & 0.764     & 0.800&-&-&-&-&-&-&-&-&- \\
     OEF~\cite{hallman2015oef}      &-& CVPR'15      & 0.746     & 0.770     & 0.815&-&-&-&-&-&-&-&-&- \\
    \toprule
     DeepEdge~\cite{bertasius2015deepedge} & AlexNet
     %~\cite{krizhevsky2017imagenet}
     & CVPR'15      & 0.753     & 0.772     & 0.807 &-&-&-&-&-&-&-&-&-\\
    DeepContour~\cite{shen2015deepcontour}&AlexNet  & CVPR'15      & 0.757     & 0.776     & 0.790 &-&-&-&-&-&-&-&-&-\\
    HED~\cite{xie2015holistically}   &VGG16  & ICCV'15      & 0.788     & 0.808     & 0.840&-&-&-&-&-&-&-&-&- \\
    Deep Boundary~\cite{kokkinos2015pushing}&VGG16 &ICLR'15   &0.789&0.811&0.789&0.803&0.820&0.848  &0.809&0.827&0.861 & 0.813     & 0.831     & 0.866 \\
    CEDN~\cite{yang2016object}     &VGG16  &  CVPR'16      & 0.788     & 0.804     & -&-&-&-&-&-&-&-&-&- \\
    RDS~\cite{liu2016learning}   &VGG16 & CVPR'16      & 0.792     & 0.810     & 0.818 &-&-&-&-&-&-&-&-&-\\
    COB~\cite{maninis2016cob}         &    VGG16  & ECCV'16      & 0.793     & 0.820     & 0.859&-&-&-&-&-&-&-&-&- \\
    AMH-Net~\cite{xu2017AMHNet}    &ResNet50 & NeurIPS'17   & 0.798     & 0.829     & 0.869&-&-&-&-&-&-&-&-&- \\
    RCF~\cite{liu2017richer}    & VGG16  & CVPR'17   &0.798&0.815&-&-&-& -  & 0.806     & 0.823     & - &0.811     & 0.830     & 0.846\\
    CED~\cite{wang2017ced}   &VGG16   & CVPR'17   &0.803&0.820&0.871&-&-&-   & 0.815     & 0.833     & \SBEST{0.889} &-&-&-\\
    LPCB~\cite{deng2018learning} &VGG16 & ECCV'18    &0.800&0.816&-&-&-&-&0.808&0.824&-  & 0.815     & 0.834     & - \\
    BDCN~\cite{he2019bi} &VGG16 & CVPR'19   &0.806&0.826&0.847&-&-&-&0.820&0.838&0.888   & 0.828     & 0.844     & 0.890 \\
     DSCD~\cite{deng2020deep}  & VGG16 & ACMMM'20 &0.802&0.817&-&-&-&-&0.813&0.836&-    & 0.822     &{0.859}     & - \\
    LDC~\cite{deng2021learning}    & MobileNetV2
    & ACMMM'21   &0.799&0.816&0.837&-&-&- &0.812&0.826&0.857 & 0.819     &0.834     & 0.860 \\
    PiDiNet~\cite{su2021pixel}  &PDC & ICCV'21     &-&-&- &-&-&-& 0.807     & 0.823     & -&-&-&-\\
     FCL-Net~\cite{xuan2022fcl}&VGG16& NN'22   &0.807&0.822&-&0.816&0.833&-&0.815&0.834&-&    0.826     & 0.845     & - \\
    \toprule
    EDTER~\cite{pu2022edter}     & Transformer &CVPR'22  & \SBEST{0.824}           & \SBEST{0.841}         & \SBEST{0.880} & \BEST{0.840}   & \BEST{0.858} & \SBEST{0.896}& \SBEST{0.832}           & \SBEST{0.847}         & 0.886 & \BEST{0.848}    & \BEST{0.865}  & \SBEST{0.903} \\
          
     \toprule
     \multicolumn{1}{c|}{\multirow{2}*{UAED (Ours)}} &VGG16&  -      &0.808& 0.827&0.872 &0.819  &0.838  &0.881  &0.820   &0.840  &\SBEST{0.889}& 0.830 &0.850 &0.895 \\
     \multicolumn{1}{c|}{~} &EfficientNet&  -      & \BEST{0.829} & \BEST{0.847} & \BEST{0.892} &\SBEST{0.837}  & \SBEST{0.855} &\BEST{0.897} &\BEST{0.838}  & \BEST{0.855} &\BEST{0.902}&  \SBEST{0.844}& \SBEST{0.864} & \BEST{0.905}\\
    \bottomrule
\end{tabular}
\label{table1}
\vspace{-10pt}
\end{table*}

\textbf{BSDS500} contains 500 RGB natural images, of which 200 are for training, 100 for validation, and 200 for testing. Each image is manually annotated by 4-9 annotators. Data augmentation follows LPCB~\cite{deng2018learning}, which rotates each image at 25 different angles and selects the largest rectangle. Then each image is flipped (horizontally, vertically, and a combination of both) at each angle. So the scale of the training dataset has expanded by 100 times. Moreover, PASCAL VOC Context Dataset~\cite{everingham2010pascal} with 10,103 images is used as the additional training data, whose edge annotations are obtained from the semantic masks by the Laplacian detector.

\textbf{Multicue} is composed of 100 scenes captured to study boundary and edge detection in challenging natural scenes. Each scene contains a left-view and a right-view short (10-frame) sequence, and the last frame of each left-view sequence is labeled with edges by six annotators and boundaries by five annotators. Data is augmented by rotating at four different angles (0, 90, 180, 270) and flipping. 80 images are randomly selected for training and the remaining 20 images are for testing. This process is repeated three times and the average scores of three independent trials 
are regarded as the final results. 

\subsection{Implementation Details}
We use Pytorch~\cite{Paszke2017automatic27} based image segmentation (SMP) neural network library~\cite{Iakubovskii:2019} as the deep learning framework to implement UAED. All parameters are updated by Adam optimizer~\cite{Kingma2014Adam28}. Our model is trained with batchsize 4. The weight decay is set to 5e-4, and the learning rate is 1e-4. To speed up the training process, we follow LPCB~\cite{deng2018learning} to make all training samples the same size, so that we can train in a mini-batch way. {For the BSDS dataset, we rotate the images to keep the same size with $321\times481$. For the Multicue dataset, each image with size $720\times1280$ is randomly cropped to $512\times512$ patches for training.}

The experiments are conducted on a single RTX 3090, and the training time of 15 epochs is about 19 hours for the BSDS dataset and 3 hours for the  Multicue dataset. During the training process, the edge prediction is supervised by one randomly sampled annotation from the available label sets, and the uncertainty is supervised by the variance of the label sets. In the inference stage, we feed the test image into our UAED and obtain a predicted label distribution, and the final predicted edge map is generated by a stochastic sampling from the predicted distribution.

\subsection{Evaluation Metric}
In the experiment, we use the widely used metrics for measuring performance. The first one is referred to the optimal dataset scale (ODS) which employs a fixed threshold for all images in the dataset, which is also called Maximum F-measure (MF). 
The second metric is called optimal image scale (OIS) which selects the optimal threshold for each image. The third one is the Average Precision (AP). Before evaluation, following previous works~\cite{liu2017richer, pu2022edter}, the predicted edge maps are processed by non-maximum suppression, and the localization tolerance is set to 0.0075, which controls the maximum allowed distance in matches between the predicted edge results and the ground truth.

\subsection{Comparison with State-of-the-art}
In this section, we compare the performance of the proposed UAED with existing excellent edge detectors, including traditional detectors such as Canny~\cite{canny1986computational}, CNN-based detectors such as HED~\cite{xie2015holistically} and RCF~\cite{liu2017richer}, and transformer-based detector EDTER~\cite{pu2022edter}.

\begin{figure}[tbp]
\setlength{\abovecaptionskip}{0pt}
\setlength{\belowcaptionskip}{0pt}
\centering
\includegraphics[width=0.8\linewidth,height=.68\linewidth]
{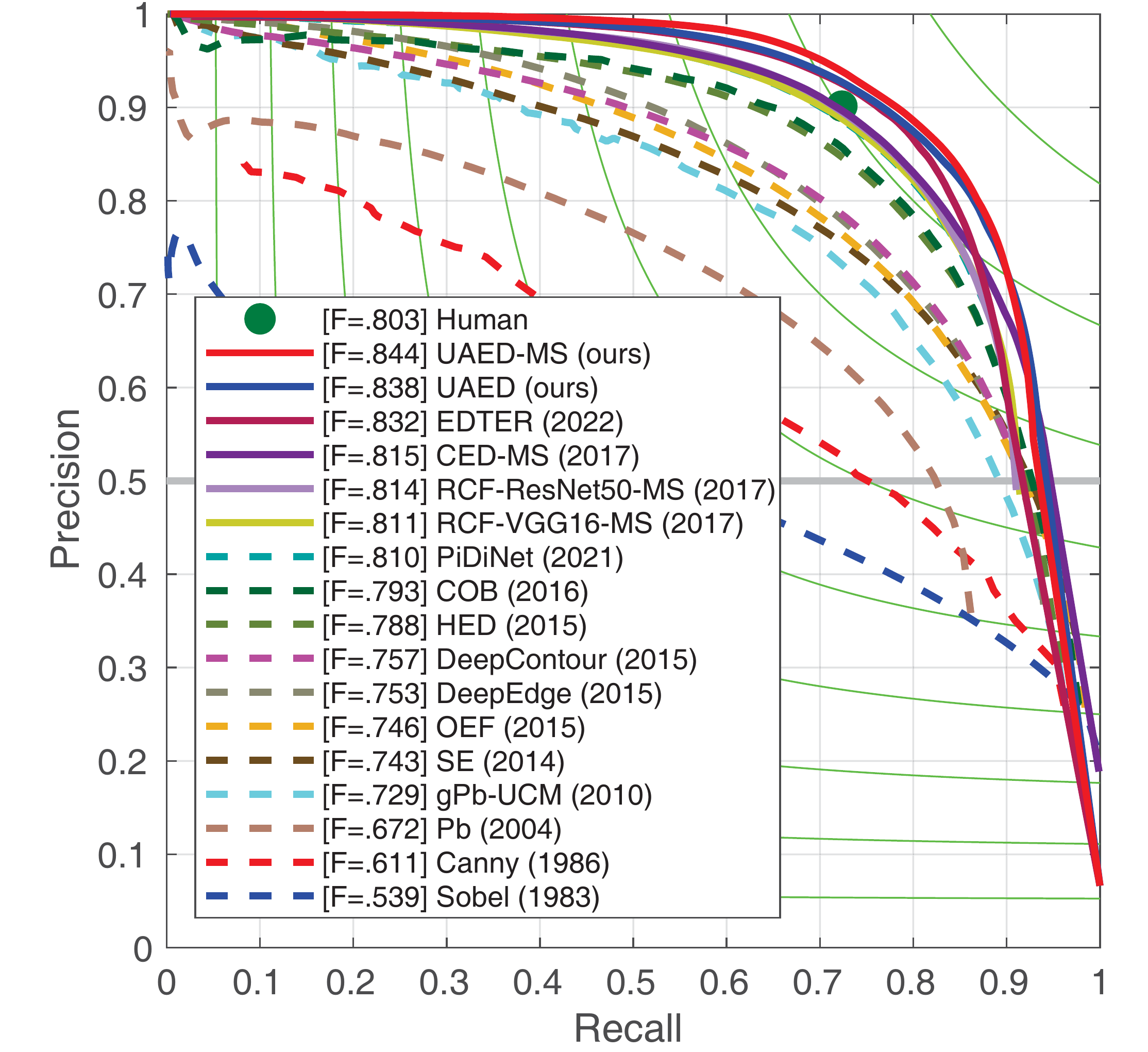}
\caption{The precision-recall curves on BSDS500.}
\label{fig3}
\vspace{-5pt}
\end{figure}

\begin{figure*}[!t]
\small
\setlength{\abovecaptionskip}{3pt}
\setlength{\belowcaptionskip}{0pt}
    \centering
	\begin{tabular}{cccccc}
	    \hspace{-.2cm}
		\includegraphics[width=.155\textwidth]{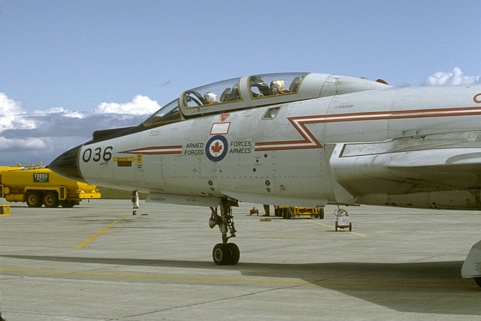} & \hspace{-.45cm}
		\includegraphics[width=.155\textwidth]{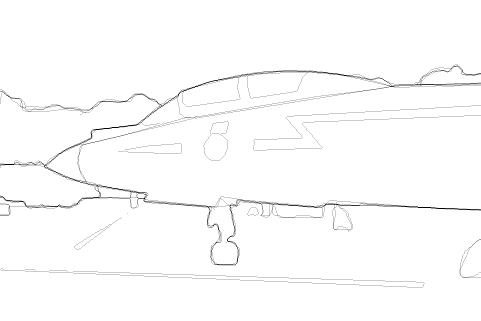} & \hspace{-.45cm}
		\includegraphics[width=.155\textwidth]{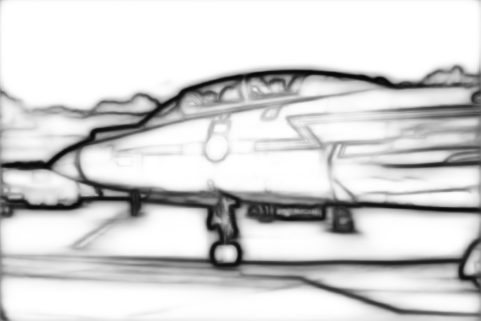}  & \hspace{-.45cm}
		\includegraphics[width=.155\textwidth]{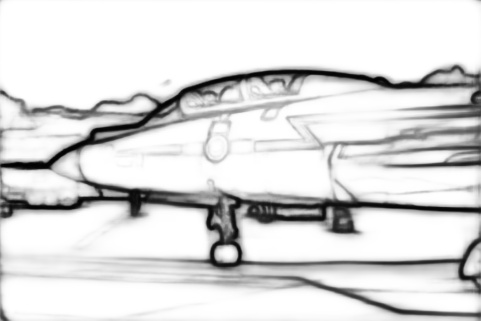} & \hspace{-.45cm}
		\includegraphics[width=.155\textwidth]{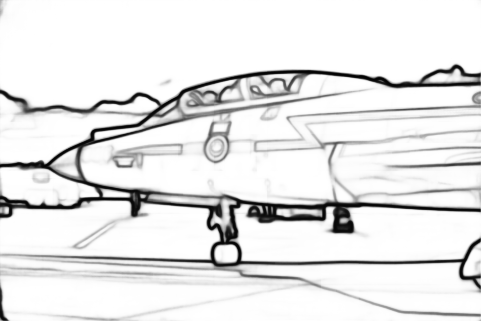}\vspace{-.06cm} & \hspace{-.45cm}
		\includegraphics[width=.155\textwidth]{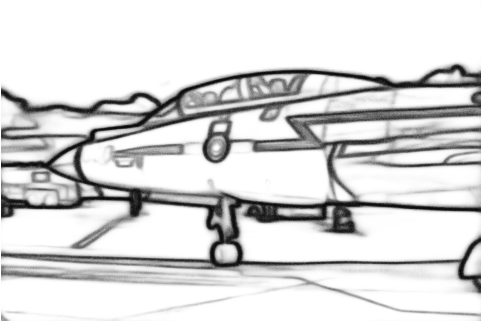}\\

		\hspace{-.2cm}
		\includegraphics[width=.155\textwidth]{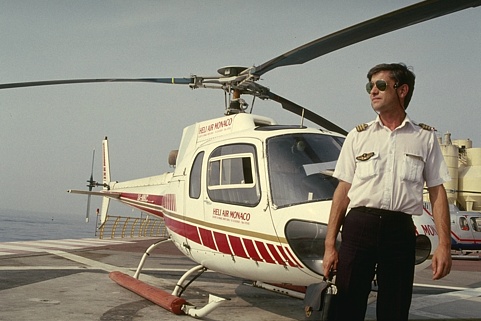} & \hspace{-.45cm}
		\includegraphics[width=.155\textwidth]{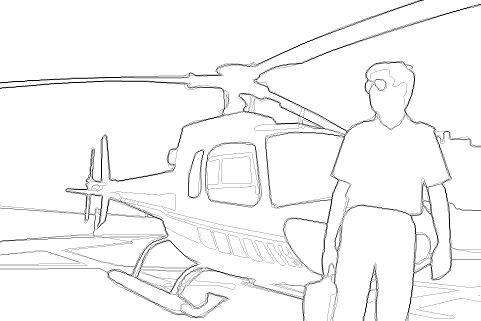} & \hspace{-.45cm}
		\includegraphics[width=.155\textwidth]{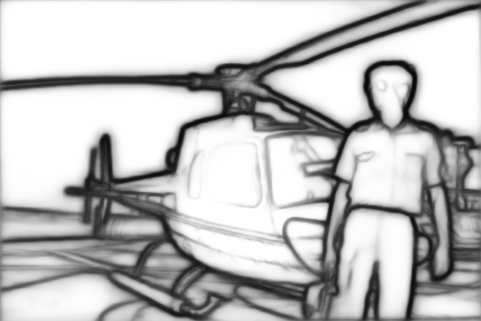}  & \hspace{-.45cm}
		\includegraphics[width=.155\textwidth]{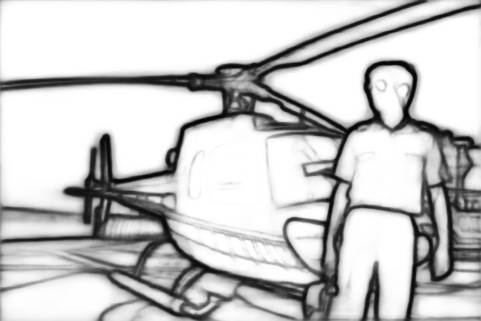} & \hspace{-.45cm}
		\includegraphics[width=.155\textwidth]{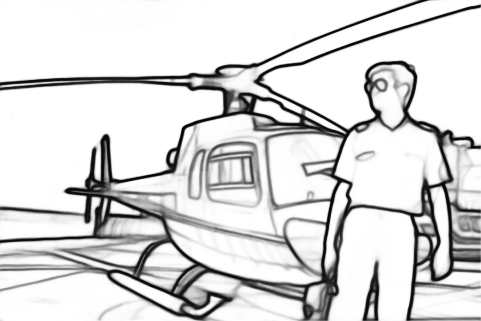}\vspace{-.06cm} & \hspace{-.45cm}
		\includegraphics[width=.155\textwidth]{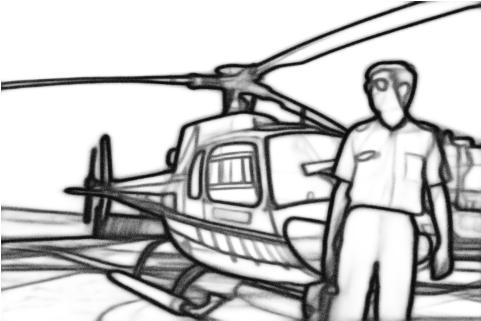}\\
		
		\hspace{-.2cm}
		\includegraphics[width=.155\textwidth]{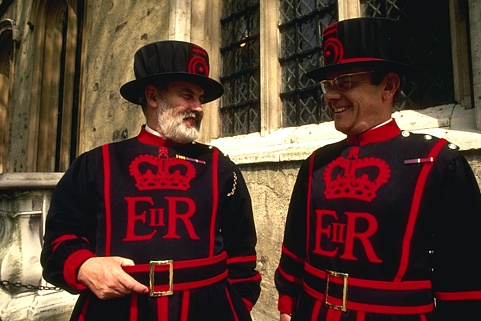} & \hspace{-.45cm}
		\includegraphics[width=.155\textwidth]{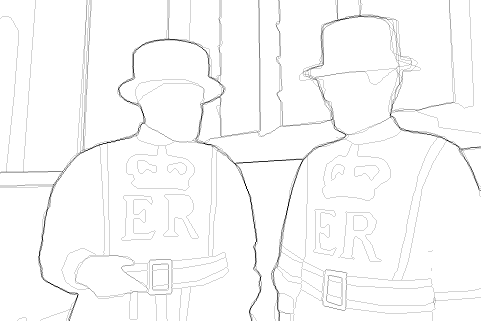} & \hspace{-.45cm}
		\includegraphics[width=.155\textwidth]{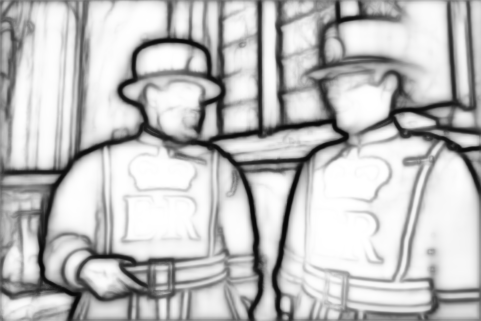} &  \hspace{-.45cm}
		\includegraphics[width=.155\textwidth]{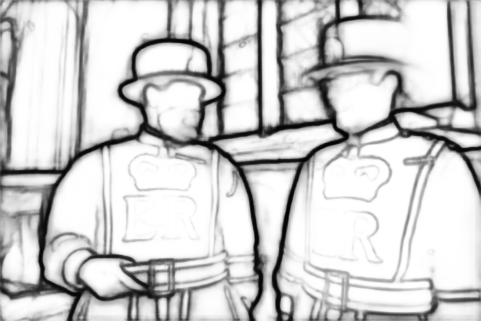} & \hspace{-.45cm}
		\includegraphics[width=.155\textwidth]{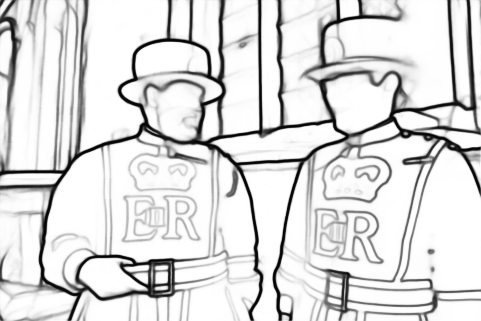}
		&\hspace{-.45cm}
		\includegraphics[width=.155\textwidth]{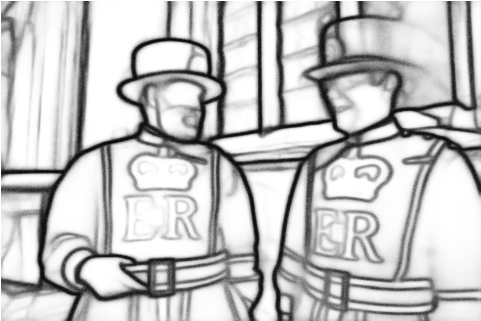} 
		\vspace{-.06cm} \\
		
		\hspace{-.2cm} (a) Input 
		& \hspace{-.45cm} (b) Ground truth 
		&\hspace{-.45cm} (c) RCF~\cite{liu2017richer} 
		&\hspace{-.45cm} (d) BDCN~\cite{he2019bi} 
		&\hspace{-.5cm} (e) EDTER~\cite{pu2022edter}  &\hspace{-.45cm} (f)  UAED(Ours) \\
	\end{tabular}
    \caption{Qualitative comparisons on three challenging samples in the BSDS500 test set.}
    \label{fig4}
    \vspace{-5pt}
\end{figure*}

\textbf{BSDS results.} The results are summarized in Table~\ref{table1}. We can see that our proposed UAED outperforms other previous CNN-based methods. In the single scale setting, compared with the second best CNN-based method BDCN~\cite{he2019bi}, we obtain a large performance gain by 2.3\%, 2.1\% and 4.5\% in terms of ODS, OIS and AP. We also achieve ODS=0.844, OIS=0.864 and AP=0.905 under the MS-VOC setting, which also surpasses BDCN~\cite{he2019bi} by a large margin (1.6\%, 2.0\% and 1.5\%). Compared with transformer-based EDTER~\cite{pu2022edter}, we still achieve the best performance and increase
the scores by 0.5\%, 0.6\%, and 1.2\% in ODS, OIS and AP metrics when testing on a single scale input. The results are only slightly lower than EDTER under multi-scale settings. The possible reason is that EDTER is learned in a patch-based manner, so the results are greatly improved when testing under the multi-scale setting. To show the results more intuitively, we give the Precision-Recall curve in Fig.~\ref{fig3}. The visualization results for some challenging samples in the BSDS testing set are shown in Fig.~\ref{fig4}. It is clear that our UAED can extract more detailed edges. {Moreover, we conduct experiments based on the VGG16~\cite{simonyan2014very} encoder since most of the CNN-based methods utilize it. The results show that we still surpass all CNN-based methods, which further verifies the effectiveness of UAED.}

\textbf{Multicue results.} Similarly, experiments are conducted on the Multicue edges and boundaries. The results are shown in Table~\ref{table2}. Our proposed UAED achieves a new state-of-the-art on the Multicue edge and boundary in three metrics (ODS=0.895, OIS=0.902, AP=0.949 in edge, and ODS=0.864, OIS=0.872, AP=0.927 in boundary). 
% For the boundary, we surpass all CNN-based methods while only perform a little worse than EDTER.

\begin{table}[t]
\setlength{\abovecaptionskip}{0pt}
\caption{Comparisons on \textbf{Multicue}~\cite{mely2016systematic}. All results are obtained by a single scale input.}
\centering
\footnotesize
\renewcommand\arraystretch{0.9}
\renewcommand\tabcolsep{1.5pt}
\begin{tabular}{c|c|c|ccc}
    \toprule
    & Method                          & Pub.'Year & ODS           & OIS           & AP            \\  
    \toprule
    \multirow{8}{*}{\rotatebox{90}{Edge}}
    & Human~\cite{mely2016systematic}   & VR'16     & .750 (0.024) & -             & -             \\
    & Multicue~\cite{mely2016systematic}& VR'16     & .830 (0.002) & -             & -             \\
    & HED~\cite{xie2015holistically}           & ICCV'15   & .851 (0.014) & .864 (0.011) & -             \\
    & RCF~\cite{liu2017richer}           & CVPR'17   & .857 (0.004) & .862 (0.004) & -             \\
    & BDCN~\cite{he2019bi}          & CVPR'19   & {.891 (0.001)} & {.898 (0.002)} & {.935(0.002)}  \\
    & DSCD~\cite{deng2020deep}        & ACMMM'20  & .871 (0.007) & .876 (0.002) & -             \\
     & LDC~\cite{deng2021learning}        & ACMMM'21  & .881 (0.012) & .893 (0.011) & -             \\
    & PiDiNet~\cite{su2021pixel}    & ICCV'21   & .855 (0.007) & .860 (0.005) & -              \\
     & FCL-Net~\cite{xuan2022fcl}        & NN'22  & .875 (0.005) & .880 (0.005) & -             \\
    & EDTER~\cite{pu2022edter}                    & CVPR'22         &\SBEST {.894 (0.005)} &\SBEST {.900 (0.003)} & \SBEST{.944 (0.002)}     \\
    & UAED (Ours)   & - &\BEST{ .895 (0.002)}  & \BEST{.902 (0.001)}&\BEST{.949 (0.002) }\\
    \toprule
    \multirow{8}{*}{\rotatebox{90}{Boundary}}
    & Human~\cite{mely2016systematic}   & VR'16     & .760 (0.017) & -             & -             \\
    & Multicue~\cite{mely2016systematic}& VR'16     & .720 (0.014) & -             & -             \\
    & HED~\cite{xie2015holistically}           & ICCV'15   & .814 (0.011) & .822 (0.008) & .869 (0.015) \\
    & RCF~\cite{liu2017richer}           & CVPR'17   & .817 (0.004) & .825 (0.005) & -             \\
    & BDCN~\cite{he2019bi}          & CVPR'19   & {.836 (0.001)} & {.846 (0.003)} & {.893 (0.001)} \\
    & DSCD~\cite{deng2020deep}        & ACMMM'20  & .828 (0.003) & .835 (0.004) & -             \\
    & LDC~\cite{deng2021learning}        & ACMMM'21  & .839 (0.012) & .853 (0.006) & -             \\
    & PiDiNet~\cite{su2021pixel}    & ICCV'21   & .818 (0.003) & .830 (0.005) & -             \\
    & FCL-Net~\cite{xuan2022fcl}        & NN'22  & .834 (0.016) & .840 (0.016) & -  \\
    & EDTER~\cite{pu2022edter}                    & CVPR'22         & \SBEST{.861 (0.003)} & \SBEST{.870 (0.004)} & \SBEST{.919 (0.003)}\\
    & UAED (Ours)                    & - &\BEST{.864 (0.004)} &\BEST{.872 (0.006)} &\BEST{.927 (0.006)} \\
    \bottomrule
  \end{tabular}
\label{table2}
\vspace{-10pt}
\end{table}
\vspace{0.1em}

\subsection{Ablation Study}\label{abstudy}
There are several strategies influencing the performance, including adding a variance branch for uncertainty estimation (UE), supervising the estimation of variance by multiple annotations ($\mathcal{L}_{\rm bvar}$), and weighting BCE loss by estimated uncertainty ($\mathcal{L}_{\rm uedge}$). We conduct ablation studies on those aspects to explore the role of each part, and the results are shown in Table~\ref{table3}. All experiments are validated on the single scale with and without PASCAL VOC pre-training.
\begin{table}[ht]\small
 \centering
 \setlength{\abovecaptionskip}{0pt}
 \caption{The ablation study on the BSDS500 dataset for the role of every part plays. The results are obtained on a single scale input.}
 \setlength{\tabcolsep}{1.2mm}{
\begin{tabular}{c|c|ccc|ccc}
    \toprule
    $\#$ &\makecell[c]{Method}&\makecell[c]{UE}&$\mathcal{L}_{\rm bvar}$  & \makecell[c]{$\mathcal{L}_{\rm uedge}$ }  &\makecell[c]{ODS}&\makecell[c]{OIS}&\makecell[c]{AP}\\
    \toprule
    1&{Baseline}&&& & 0.824 & 0.841 & 0.875\\
    \toprule
    2& \multirow{3}*{\shortstack{UAED}}&\checkmark& &  &  0.825 &0.843 &0.884\\
    
    3&& \checkmark&\checkmark & &0.829 &0.846 & 0.892\\
    
    4&& \checkmark&\checkmark  &\checkmark &0.829 &0.847 &0.892 \\
    \toprule
    5&{Baseline-VOC}&&& & 0.831 & 0.846 & 0.887\\
    \toprule
    6&\multirow{3}*{\shortstack{UAED-VOC}} &\checkmark& &  &  0.834 &0.852&0.895\\
    7&& \checkmark&\checkmark & &0.836 &0.853 &0.901 \\
    8&& \checkmark&\checkmark  &\checkmark &0.838 &0.855 &0.902 \\
    \bottomrule
    \end{tabular}}
  \label{table3}%
  \vspace{-10pt}
\end{table}
Experiments $\#1$ and $\#5$ are used as baseline methods, which fuse the multiple annotations into one single deterministic label according to a fixed threshold (0.3). The experiments
($\#2$-$\#4$ and $\#6$-$\#8$) are the results of our proposed UAED by deactivating different strategies which utilize all available multiple annotations.

\textbf{The effect of the uncertainty estimation (UE).} We first add a new decoder branch to the edge detector and convert the label space into a learnable distribution, which can be used to estimate uncertainty. Instead of generating deterministic predictions ($\#1$ and $\#5$), experiments ($\#2$ and $\#6$) regard the prediction as a distribution, and possess two independent decoders that represent mean and variance respectively. We can observe that experiment $\#2$ obtains 0.1\%, 0.2\%, and 0.9\% in ODS, OIS, and AP than experiment $\#1$. 
Experiment $\#6$ obtains 0.3\%, 0.6\%, and 0.8\% in ODS, OIS, and AP than experiment $\#5$. The improvements clearly show the effectiveness of introducing uncertainty.

\textbf{The effect of $\mathcal{L}_{\rm bvar}$.} If there is not any explicit supervision of the uncertainty, the learning process will not be easy. Fortunately, we can compute the variance from the provided label sets as the variance supervision ($\#3$ and $\#7$). From the ablations, we can see that experiment $\#3$ increases the score by 0.4\% (ODS), 0.3\% (OIS), and 0.8\% (AP) than $\#2$. Similarly, experiment $\#7$ obtains 0.2\% (ODS), 0.1\% (OIS), and 0.6\% (AP) than $\#6$. The improvements clearly show the benefit of this explicit supervision.

\begin{table}[ht]
\setlength{\abovecaptionskip}{3pt}
\setlength{\belowcaptionskip}{0pt}
 \centering
 \caption{The experiments of different weighting methods. All results are obtained by a single scale input. }
 \setlength{\tabcolsep}{2mm}{
\begin{tabular}{c|ccc}
    \toprule
    Method & ODS  & \makecell[c]{OIS} &\makecell[c]{AP}  \\
    \toprule
    $e^{-\hat{\sigma}}\mathcal{L}_{\rm edge}+2\hat{\sigma}$& 0.825 &0.843  &0.884 \\
    $e^{\hat{\sigma}}\mathcal{L}_{\rm edge}$& 0.826 &  0.846& 0.891  \\
    $e^{ \beta_t  \sigma} \mathcal{L}_{\rm edge}$& 0.826 &0.845  &0.890\\
    \toprule
    $e^{ \beta_t  \hat{\sigma}} \mathcal{L}_{\rm edge}$&  0.829&0.847  &0.892\\
    \bottomrule
    \end{tabular}}
  \label{table4}%
   \vspace{-10pt}
\end{table}

\textbf{The effect of $\mathcal{L}_{\rm uedge}$.} The progressive weighting loss makes the network first learn the variance as uncertainty, then utilize the learned uncertainty to progressively focus on the pixels with higher uncertainty. The experiments ($\#4$ and $\#8$) show that the weighting strategy can further improve  performance.

\textbf{The effect of different weighting loss.}
Instead of the traditional strategy (lower weight for higher uncertainty), we design a progressive weighting loss to emphasize the pixels with higher uncertainty. To verify this, we conduct ablation experiments shown in Table~\ref{table4}. Compared with other strategies, \ie the traditional weighting loss (the first row), the fixed weighting loss (the second row), and the ground-truth label variance for weighting loss (the third row), our designed strategy achieves the best performance. 
%in three metrics.

\begin{table}[ht]\small
 \centering
\setlength{\abovecaptionskip}{0pt}
\caption{The experiments on the BSDS500 dataset for different uncertainty estimation (UE) strategies. All results are obtained by a single scale input.}
 \setlength{\tabcolsep}{1.8mm}{
\begin{tabular}{c|c|ccc}
    \toprule
    UE Type &UE Method & ODS  & \makecell[c]{OIS} &\makecell[c]{AP} \\
    \toprule
   -& Baseline& 0.824 & 0.841 & 0.875  \\
   \toprule
    \multirow{2}*{\shortstack{Epistemic\\(Model)}}&MC Dropout& 0.825&0.842  &0.882  \\
    &RBUE & 0.825  & 0.843 &0.882  \\
    \toprule
     \multirow{3}*{\shortstack{Aleatoric\\(Data)}}&CVAE-based& 0.821 &0.844   &0.880  \\
    &EBM-based & 0.824& 0.842 & 0.884\\
    &Probabilistic Embedding & 0.823& 0.839 &0.885 \\
    \toprule
    Data&UAED (Ours)&0.829&0.847& 0.892 \\
    \bottomrule
    \end{tabular}}
  \label{table5}%
  \vspace{-5pt}
\end{table}

\setlength\abovedisplayskip{5pt}
\subsection{Further Analysis} \label{arch}
{\textbf{Comparison with different uncertainty estimation methods.}} Uncertainty estimation is the most vital part of UAED, so we explore the influence of different methods. The popular uncertainty estimation methods include MC dropout~\cite{gal2016dropout}, RBUE~\cite{xia2021rbue}, CVAE-based~\cite{Zhang2020UCnet13}, EBM-based~\cite{du2019implicit}, and probabilistic embedding~\cite{shi2019probabilistic}.  MC Dropout~\cite{gal2016dropout} and RBUE~\cite{xia2021rbue} model epistemic (model) uncertainty, {and both of them usually require} hundreds of forwards to obtain excellent results. Generative model based methods, including CVAE-based and EBM-based models, learn low-level latent space, which captures randomness caused by the data.
\begin{table}[ht]\small
 \centering
 \setlength{\abovecaptionskip}{0pt}
 \caption{The experiments about different encoder and decoder structures. All results are obtained by a single scale input.}
 \setlength{\tabcolsep}{0.4mm}{
\begin{tabular}{c|lll}
    \toprule
    Encoder / Decoder & ODS  & \makecell[l]{OIS} &\makecell[l]{AP}  \\
    \toprule
    VGG~\cite{simonyan2014very} / UNet++~\cite{zhou2018unet++}& 0.801 &0.820  &0.862 \\
   {+UAED} & 0.808(\textcolor{red}{\scriptsize{+.007}}) &  0.827(\textcolor{red}{\scriptsize{+.007}})& 0.872(\textcolor{red}{\scriptsize{+.010}})  \\
   
    \toprule
    EfficientNet~\cite{tan2019efficientnet}/UNet\cite{ronneberger2015u}&  0.821&0.836   &0.878  \\
    {+UAED} &0.824(\textcolor{red}{\scriptsize{+.003}})&0.843(\textcolor{red}{\scriptsize{+.007}})  &0.886(\textcolor{red}{\scriptsize{+.008}}) \\
    \toprule
    SegFormer~\cite{xie2021segformer} / UNet~\cite{ronneberger2015u}&  0.822&0.838  &0.873\\
   {+UAED} &  0.828(\textcolor{red}{\scriptsize{+.006}})&0.845(\textcolor{red}{\scriptsize{+.007}})  &0.888(\textcolor{red}{\scriptsize{+.015}}) \\
    \bottomrule
    \end{tabular}}
  \label{table6}%
   \vspace{-10pt}
\end{table}
Probabilistic embedding builds distribution in the feature space while the proposed UAED differs in the label distribution, and both model aleatoric (data) uncertainty. The details of different uncertainty estimation methods can be found in the supplementary material.

Table~\ref{table5} presents the comparison. MC dropout and RBUE only bring minor improvement and other methods are even inferior to baseline. By introducing the uncertainty to the label space, UAED obtains the best improvement.

{\textbf{The improvement for different backbones.}} Our proposed UAED is a play-and-plug module, so we combine it with diverse encoder-decoder architectures to verify its effectiveness. Specifically, the encoder contains CNN-based VGG~\cite{simonyan2014very}, EfficientNet~\cite{tan2019efficientnet}, and transformer-based SegFormer~\cite{xie2021segformer}. The decoder includes popular UNet~\cite{ronneberger2015u} and UNet++~\cite{zhou2018unet++}. All encoders are pre-trained on the ImageNet~\cite{deng2009imagenet} dataset. The results in Table~\ref{table6} show a
consistent performance gain of the proposed UAED, validating
that UAED can be easily combined with the existing frameworks to boost performance consistently.

{\textbf{Computational cost.}
Our experiments are conducted on a single RTX 3090, %with 12G GPU memory, 
which consume 12G GPU memory for the BSDS dataset with a batch size of 4. We also compute the inference time. The speed of our proposed UAED is 17 FPS, which increases not
too much (19 FPS for the encoder-decoder baseline model). By comparison, the transformer-based EDTER~\cite{pu2022edter} consumes 15G GPU (stage I) and 14G (stage II) with batchsize of 1 for training and runs at 5 FPS for inference. The statistics indicate that our proposed UAED obtains a comparable performance only using relatively less time and resources.}
\section{Conclusion}
We are the first work to employ uncertainty into the edge detector to model the inherent ambiguity underlying multiple annotations. The uncertainty-aware edge detector (UAED) regards the label space as the distribution and adds a branch to estimate the variance, which can be further utilized to progressively weight the optimized loss function. We conduct experiments on BSDS and Multicue datasets. The results demonstrate that our proposed UAED can bring consistent improvement by exploring the uncertainty beneath the multiple annotations.

\vspace{0.3em} 
\noindent\textbf{Limitation.} This method still needs labor-consuming pixel-level annotations. How to utilize fewer annotations to achieve competitive results remains an open issue.

\vspace{0.3em}
\noindent\textbf{Acknowledgements.} This work is supported by National Natural Science Foundation of China (62271042) and Beijing Natural Science Foundation (M22022, L211015).

{\small
\bibliographystyle{ieee_fullname}
\bibliography{egbib}

\begin{thebibliography}{10}\itemsep=-1pt

\bibitem{arbelaez2010contour}
Pablo Arbelaez, Michael Maire, Charless Fowlkes, and Jitendra Malik.
\newblock Contour detection and hierarchical image segmentation.
\newblock {\em IEEE Trans. Pattern Anal. Mach. Intell.}, 33(5):898--916, 2010.

\bibitem{Baumgartner2019PhiSeg21}
Christian~F Baumgartner, Kerem~C Tezcan, Krishna Chaitanya, Andreas~M
  H{\"o}tker, Urs~J Muehlematter, Khoschy Schawkat, Anton~S Becker, Olivio
  Donati, and Ender Konukoglu.
\newblock Phiseg: Capturing uncertainty in medical image segmentation.
\newblock In {\em International Conference on Medical Image Computing and
  Computer-Assisted Intervention}, pages 119--127. Springer, 2019.

\bibitem{bertasius2015deepedge}
Gedas Bertasius, Jianbo Shi, and Lorenzo Torresani.
\newblock Deepedge: A multi-scale bifurcated deep network for top-down contour
  detection.
\newblock In {\em IEEE Conf. Comput. Vis. Pattern Recog.}, pages 4380--4389,
  2015.

\bibitem{canny1986computational}
John Canny.
\newblock A computational approach to edge detection.
\newblock {\em IEEE Trans. Pattern Anal. Mach. Intell.}, (6):679--698, 1986.

\bibitem{chang2020data}
Jie Chang, Zhonghao Lan, Changmao Cheng, and Yichen Wei.
\newblock Data uncertainty learning in face recognition.
\newblock In {\em IEEE Conf. Comput. Vis. Pattern Recog.}, pages 5710--5719,
  2020.

\bibitem{deng2009imagenet}
Jia Deng, Wei Dong, Richard Socher, Li-Jia Li, Kai Li, and Li Fei-Fei.
\newblock Imagenet: A large-scale hierarchical image database.
\newblock In {\em IEEE Conf. Comput. Vis. Pattern Recog.}, pages 248--255.
  Ieee, 2009.

\bibitem{deng2020deep}
Ruoxi Deng and Shengjun Liu.
\newblock Deep structural contour detection.
\newblock In {\em ACM Int. Conf. Multimedia}, pages 304--312, 2020.

\bibitem{deng2021learning}
Ruoxi Deng, Shengjun Liu, Jinxin Wang, Huibing Wang, Hanli Zhao, and Xiaoqin
  Zhang.
\newblock Learning to decode contextual information for efficient contour
  detection.
\newblock In {\em ACM Int. Conf. Multimedia}, pages 4435--4443, 2021.

\bibitem{deng2018learning}
Ruoxi Deng, Chunhua Shen, Shengjun Liu, Huibing Wang, and Xinru Liu.
\newblock Learning to predict crisp boundaries.
\newblock In {\em Eur. Conf. Comput. Vis.}, pages 562--578, 2018.

\bibitem{dollar2014fast}
Piotr Doll{\'a}r and C~Lawrence Zitnick.
\newblock Fast edge detection using structured forests.
\newblock {\em IEEE Trans. Pattern Anal. Mach. Intell.}, 37(8):1558--1570,
  2014.

\bibitem{du2019implicit}
Yilun Du and Igor Mordatch.
\newblock Implicit generation and generalization in energy-based models.
\newblock {\em arXiv preprint arXiv:1903.08689}, 2019.

\bibitem{everingham2010pascal}
Mark Everingham, Luc Van~Gool, Christopher~KI Williams, John Winn, and Andrew
  Zisserman.
\newblock The pascal visual object classes (voc) challenge.
\newblock {\em Int. J. Comput. Vis.}, 88(2):303--338, 2010.

\bibitem{gal2016dropout}
Yarin Gal and Zoubin Ghahramani.
\newblock Dropout as a bayesian approximation: Representing model uncertainty
  in deep learning.
\newblock In {\em international conference on machine learning}, pages
  1050--1059. PMLR, 2016.

\bibitem{gustafsson2020energy}
Fredrik~K Gustafsson, Martin Danelljan, Goutam Bhat, and Thomas~B Sch{\"o}n.
\newblock Energy-based models for deep probabilistic regression.
\newblock In {\em Eur. Conf. Comput. Vis.}, pages 325--343. Springer, 2020.

\bibitem{hallman2015oef}
Sam Hallman and Charless~C Fowlkes.
\newblock Oriented edge forests for boundary detection.
\newblock In {\em IEEE Conf. Comput. Vis. Pattern Recog.}, pages 1732--1740,
  2015.

\bibitem{he2019bi}
Jianzhong He, Shiliang Zhang, Ming Yang, Yanhu Shan, and Tiejun Huang.
\newblock Bi-directional cascade network for perceptual edge detection.
\newblock In {\em IEEE Conf. Comput. Vis. Pattern Recog.}, pages 3828--3837,
  2019.

\bibitem{he2016deep}
Kaiming He, Xiangyu Zhang, Shaoqing Ren, and Jian Sun.
\newblock Deep residual learning for image recognition.
\newblock In {\em IEEE Conf. Comput. Vis. Pattern Recog.}, pages 770--778,
  2016.

\bibitem{howard2018inverted}
Andrew Howard, Andrey Zhmoginov, Liang-Chieh Chen, Mark Sandler, and Menglong
  Zhu.
\newblock Inverted residuals and linear bottlenecks: Mobile networks for
  classification, detection and segmentation.
\newblock 2018.

\bibitem{hwang2014contour}
Jyh-Jing Hwang and Tyng-Luh Liu.
\newblock Contour detection using cost-sensitive convolutional neural networks.
\newblock {\em arXiv preprint arXiv:1412.6857}, 2014.

\bibitem{Iakubovskii:2019}
Pavel Iakubovskii.
\newblock Segmentation models pytorch.
\newblock \url{https://github.com/qubvel/segmentation_models.pytorch}, 2019.

\bibitem{iandola2016squeezenet}
Forrest~N Iandola, Song Han, Matthew~W Moskewicz, Khalid Ashraf, William~J
  Dally, and Kurt Keutzer.
\newblock Squeezenet: Alexnet-level accuracy with 50x fewer parameters and< 0.5
  mb model size.
\newblock {\em arXiv preprint arXiv:1602.07360}, 2016.

\bibitem{ioffe2015batch}
Sergey Ioffe and Christian Szegedy.
\newblock Batch normalization: Accelerating deep network training by reducing
  internal covariate shift.
\newblock In {\em International conference on machine learning}, pages
  448--456. PMLR, 2015.

\bibitem{kendall2017uncertainties}
Alex Kendall and Yarin Gal.
\newblock What uncertainties do we need in bayesian deep learning for computer
  vision?
\newblock {\em Adv. Neural Inform. Process. Syst.}, 30, 2017.

\bibitem{Kingma2014Adam28}
Diederik~P Kingma and Jimmy Ba.
\newblock Adam: A method for stochastic optimization.
\newblock {\em arXiv preprint arXiv:1412.6980}, 2014.

\bibitem{kingma2015variational}
Durk~P Kingma, Tim Salimans, and Max Welling.
\newblock Variational dropout and the local reparameterization trick.
\newblock {\em Adv. Neural Inform. Process. Syst.}, 28, 2015.

\bibitem{kittler1983accuracy}
Josef Kittler.
\newblock On the accuracy of the sobel edge detector.
\newblock {\em Image and Vision Computing}, 1(1):37--42, 1983.

\bibitem{Kohl2018probabilistic20}
Simon Kohl, Bernardino Romera-Paredes, Clemens Meyer, Jeffrey De~Fauw, Joseph~R
  Ledsam, Klaus Maier-Hein, SM Eslami, Danilo Jimenez~Rezende, and Olaf
  Ronneberger.
\newblock A probabilistic u-net for segmentation of ambiguous images.
\newblock {\em Adv. Neural Inform. Process. Syst.}, 31, 2018.

\bibitem{kokkinos2015pushing}
Iasonas Kokkinos.
\newblock Pushing the boundaries of boundary detection using deep learning.
\newblock {\em Int. Conf. Learn. Represent.}, 2016.

\bibitem{kumar2019videoflow}
Manoj Kumar, Mohammad Babaeizadeh, Dumitru Erhan, Chelsea Finn, Sergey Levine,
  Laurent Dinh, and Durk Kingma.
\newblock Videoflow: A conditional flow-based model for stochastic video
  generation.
\newblock {\em arXiv preprint arXiv:1903.01434}, 2019.

\bibitem{lakshminarayanan2017simple}
Balaji Lakshminarayanan, Alexander Pritzel, and Charles Blundell.
\newblock Simple and scalable predictive uncertainty estimation using deep
  ensembles.
\newblock {\em Adv. Neural Inform. Process. Syst.}, 30, 2017.

\bibitem{liu2020dynamic}
Jiang-Jiang Liu, Qibin Hou, and Ming-Ming Cheng.
\newblock Dynamic feature integration for simultaneous detection of salient
  object, edge, and skeleton.
\newblock {\em IEEE Trans. Image Process.}, 29:8652--8667, 2020.

\bibitem{liu2017richer}
Yun Liu, Ming-Ming Cheng, Xiaowei Hu, Kai Wang, and Xiang Bai.
\newblock Richer convolutional features for edge detection.
\newblock In {\em IEEE Conf. Comput. Vis. Pattern Recog.}, pages 3000--3009,
  2017.

\bibitem{liu2016learning}
Yu Liu and Michael~S Lew.
\newblock Learning relaxed deep supervision for better edge detection.
\newblock In {\em IEEE Conf. Comput. Vis. Pattern Recog.}, pages 231--240,
  2016.

\bibitem{maninis2016cob}
Kevis-Kokitsi Maninis, Jordi Pont-Tuset, Pablo Arbel{\'a}ez, and Luc Van~Gool.
\newblock Convolutional oriented boundaries.
\newblock In {\em Eur. Conf. Comput. Vis.}, pages 580--596. Springer, 2016.

\bibitem{martin2004learning}
David~R Martin, Charless~C Fowlkes, and Jitendra Malik.
\newblock Learning to detect natural image boundaries using local brightness,
  color, and texture cues.
\newblock {\em IEEE Trans. Pattern Anal. Mach. Intell.}, 26(5):530--549, 2004.

\bibitem{mely2016systematic}
David~A M{\'e}ly, Junkyung Kim, Mason McGill, Yuliang Guo, and Thomas Serre.
\newblock A systematic comparison between visual cues for boundary detection.
\newblock {\em Vision research}, 120:93--107, 2016.

\bibitem{nazeri2019edgeconnect}
Kamyar Nazeri, Eric Ng, Tony Joseph, Faisal Qureshi, and Mehran Ebrahimi.
\newblock Edgeconnect: Structure guided image inpainting using edge prediction.
\newblock In {\em IEEE Conf. Comput. Vis. Pattern Recog. Worksh.}, pages 0--0,
  2019.

\bibitem{Paszke2017automatic27}
Adam Paszke, Sam Gross, Soumith Chintala, Gregory Chanan, Edward Yang, Zachary
  DeVito, Zeming Lin, Alban Desmaison, Luca Antiga, and Adam Lerer.
\newblock Automatic differentiation in pytorch.
\newblock 2017.

\bibitem{patro2019u}
Badri~N Patro, Mayank Lunayach, Shivansh Patel, and Vinay~P Namboodiri.
\newblock U-cam: Visual explanation using uncertainty based class activation
  maps.
\newblock In {\em Int. Conf. Comput. Vis.}, pages 7444--7453, 2019.

\bibitem{pu2021rindnet}
Mengyang Pu, Yaping Huang, Qingji Guan, and Haibin Ling.
\newblock Rindnet: Edge detection for discontinuity in reflectance,
  illumination, normal and depth.
\newblock In {\em Int. Conf. Comput. Vis.}, pages 6879--6888, 2021.

\bibitem{pu2022edter}
Mengyang Pu, Yaping Huang, Yuming Liu, Qingji Guan, and Haibin Ling.
\newblock Edter: Edge detection with transformer.
\newblock In {\em IEEE Conf. Comput. Vis. Pattern Recog.}, pages 1402--1412,
  2022.

\bibitem{qin2019basnet}
Xuebin Qin, Zichen Zhang, Chenyang Huang, Chao Gao, Masood Dehghan, and Martin
  Jagersand.
\newblock Basnet: Boundary-aware salient object detection.
\newblock In {\em IEEE Conf. Comput. Vis. Pattern Recog.}, pages 7479--7489,
  2019.

\bibitem{radosavovic2020designing}
Ilija Radosavovic, Raj~Prateek Kosaraju, Ross Girshick, Kaiming He, and Piotr
  Doll{\'a}r.
\newblock Designing network design spaces.
\newblock In {\em IEEE Conf. Comput. Vis. Pattern Recog.}, pages 10428--10436,
  2020.

\bibitem{ramachandran2017swish}
Prajit Ramachandran, Barret Zoph, and Quoc~V Le.
\newblock Swish: a self-gated activation function.
\newblock {\em arXiv preprint arXiv:1710.05941}, 7(1):5, 2017.

\bibitem{ronneberger2015u}
Olaf Ronneberger, Philipp Fischer, and Thomas Brox.
\newblock U-net: Convolutional networks for biomedical image segmentation.
\newblock In {\em International Conference on Medical image computing and
  computer-assisted intervention}, pages 234--241. Springer, 2015.

\bibitem{selvan2020uncertainty}
Raghavendra Selvan, Frederik Faye, Jon Middleton, and Akshay Pai.
\newblock Uncertainty quantification in medical image segmentation with
  normalizing flows.
\newblock In {\em International Workshop on Machine Learning in Medical
  Imaging}, pages 80--90. Springer, 2020.

\bibitem{shen2015deepcontour}
Wei Shen, Xinggang Wang, Yan Wang, Xiang Bai, and Zhijiang Zhang.
\newblock Deepcontour: A deep convolutional feature learned by positive-sharing
  loss for contour detection.
\newblock In {\em IEEE Conf. Comput. Vis. Pattern Recog.}, pages 3982--3991,
  2015.

\bibitem{shi2019probabilistic}
Yichun Shi and Anil~K Jain.
\newblock Probabilistic face embeddings.
\newblock In {\em Int. Conf. Comput. Vis.}, pages 6902--6911, 2019.

\bibitem{simonyan2014very}
Karen Simonyan and Andrew Zisserman.
\newblock Very deep convolutional networks for large-scale image recognition.
\newblock {\em arXiv preprint arXiv:1409.1556}, 2014.

\bibitem{song2018edgestereo}
Xiao Song, Xu Zhao, Hanwen Hu, and Liangji Fang.
\newblock Edgestereo: A context integrated residual pyramid network for stereo
  matching.
\newblock In {\em ACCV}, pages 20--35. Springer, 2018.

\bibitem{su2021pixel}
Zhuo Su, Wenzhe Liu, Zitong Yu, Dewen Hu, Qing Liao, Qi Tian, Matti
  Pietik{\"a}inen, and Li Liu.
\newblock Pixel difference networks for efficient edge detection.
\newblock pages 5117--5127, 2021.

\bibitem{tan2019efficientnet}
Mingxing Tan and Quoc Le.
\newblock Efficientnet: Rethinking model scaling for convolutional neural
  networks.
\newblock In {\em International conference on machine learning}, pages
  6105--6114. PMLR, 2019.

\bibitem{valiuddin2021improving}
MM Valiuddin, Christiaan~GA Viviers, Ruud~JG van Sloun, Fons van~der Sommen,
  et~al.
\newblock Improving aleatoric uncertainty quantification in multi-annotated
  medical image segmentation with normalizing flows.
\newblock In {\em Uncertainty for Safe Utilization of Machine Learning in
  Medical Imaging, and Perinatal Imaging, Placental and Preterm Image
  Analysis}, pages 75--88. Springer, 2021.

\bibitem{wang2017ced}
Yupei Wang, Xin Zhao, and Kaiqi Huang.
\newblock Deep crisp boundaries.
\newblock In {\em IEEE Conf. Comput. Vis. Pattern Recog.}, pages 3892--3900,
  2017.

\bibitem{wehrbein2021probabilistic}
Tom Wehrbein, Marco Rudolph, Bodo Rosenhahn, and Bastian Wandt.
\newblock Probabilistic monocular 3d human pose estimation with normalizing
  flows.
\newblock In {\em Int. Conf. Comput. Vis.}, pages 11199--11208, 2021.

\bibitem{xia2021rbue}
Yufeng Xia, Jun Zhang, Zhiqiang Gong, Tingsong Jiang, and Wen Yao.
\newblock Rbue: A relu-based uncertainty estimation method of deep neural
  networks.
\newblock {\em arXiv preprint arXiv:2107.07197}, 2021.

\bibitem{xiaofeng2012scg}
Ren Xiaofeng and Liefeng Bo.
\newblock Discriminatively trained sparse code gradients for contour detection.
\newblock {\em Adv. Neural Inform. Process. Syst.}, 25, 2012.

\bibitem{xie2021segformer}
Enze Xie, Wenhai Wang, Zhiding Yu, Anima Anandkumar, Jose~M Alvarez, and Ping
  Luo.
\newblock Segformer: Simple and efficient design for semantic segmentation with
  transformers.
\newblock {\em Adv. Neural Inform. Process. Syst.}, 34:12077--12090, 2021.

\bibitem{xie2015holistically}
Saining Xie and Zhuowen Tu.
\newblock Holistically-nested edge detection.
\newblock In {\em Int. Conf. Comput. Vis.}, pages 1395--1403, 2015.

\bibitem{xu2017AMHNet}
Dan Xu, Wanli Ouyang, Xavier Alameda-Pineda, Elisa Ricci, Xiaogang Wang, and
  Nicu Sebe.
\newblock Learning deep structured multi-scale features using attention-gated
  crfs for contour prediction.
\newblock In {\em Adv. Neural Inform. Process. Syst.}, pages 3961--3970, 2017.

\bibitem{xuan2022fcl}
Wenjie Xuan, Shaoli Huang, Juhua Liu, and Bo Du.
\newblock Fcl-net: Towards accurate edge detection via fine-scale corrective
  learning.
\newblock {\em Neural Networks}, 145:248--259, 2022.

\bibitem{yang2016object}
Jimei Yang, Brian Price, Scott Cohen, Honglak Lee, and Ming-Hsuan Yang.
\newblock Object contour detection with a fully convolutional encoder-decoder
  network.
\newblock In {\em IEEE Conf. Comput. Vis. Pattern Recog.}, pages 193--202,
  2016.

\bibitem{yu2021coupled}
Zhiding Yu, Rui Huang, Wonmin Byeon, Sifei Liu, Guilin Liu, Thomas Breuel,
  Anima Anandkumar, and Jan Kautz.
\newblock Coupled segmentation and edge learning via dynamic graph propagation.
\newblock {\em Adv. Neural Inform. Process. Syst.}, 34:4919--4932, 2021.

\bibitem{zhang2021dense}
Jing Zhang, Yuchao Dai, Mochu Xiang, Deng-Ping Fan, Peyman Moghadam, Mingyi He,
  Christian Walder, Kaihao Zhang, Mehrtash Harandi, and Nick Barnes.
\newblock Dense uncertainty estimation.
\newblock {\em arXiv preprint arXiv:2110.06427}, 2021.

\bibitem{Zhang2020UCnet13}
Jing Zhang, Deng-Ping Fan, Yuchao Dai, Saeed Anwar, Fatemeh~Sadat Saleh, Tong
  Zhang, and Nick Barnes.
\newblock Uc-net: Uncertainty inspired rgb-d saliency detection via conditional
  variational autoencoders.
\newblock In {\em IEEE Conf. Comput. Vis. Pattern Recog.}, pages 8582--8591,
  2020.

\bibitem{zhang2021learning}
Jing Zhang, Jianwen Xie, Nick Barnes, and Ping Li.
\newblock Learning generative vision transformer with energy-based latent space
  for saliency prediction.
\newblock {\em Adv. Neural Inform. Process. Syst.}, 34:15448--15463, 2021.

\bibitem{zhou2018unet++}
Zongwei Zhou, Md~Mahfuzur Rahman~Siddiquee, Nima Tajbakhsh, and Jianming Liang.
\newblock Unet++: A nested u-net architecture for medical image segmentation.
\newblock In {\em Deep learning in medical image analysis and multimodal
  learning for clinical decision support}, pages 3--11. Springer, 2018.

\end{thebibliography}
}

\end{document}